%% file: main.tex
\documentclass[11pt]{article}

\usepackage[]{acl}

\usepackage{times}
\usepackage{latexsym}

\usepackage[T1]{fontenc}
\usepackage[utf8]{inputenc}

\usepackage{microtype}

\usepackage{inconsolata}

\usepackage{graphicx}
\usepackage{enumitem}
\title{The Diminishing Returns of Early-Exit Decoding in Modern LLMs}

\author{
 \textbf{Rui Wei\textsuperscript{1}},
 \textbf{Rui Du\textsuperscript{1}},
 \textbf{Hanfei Yu\textsuperscript{1}},
 \textbf{Devesh Tiwari\textsuperscript{2}},
\\
 \textbf{Jian Li\textsuperscript{3}},
 \textbf{Zhaozhuo Xu\textsuperscript{1}},
 \textbf{Hao Wang\textsuperscript{1}},
\\
 \textsuperscript{1}Stevens Institute of Technology,
 \textsuperscript{2}Northeastern University,
 \textsuperscript{3}Stony Brook University,
\\
   rwei7@stevens.edu, rdu4@stevens.edu, hyu42@stevens.edu, d.tiwari@northeastern.edu,\\
   jian.li.3@stonybrook.edu, zxu79@stevens.edu, hwang9@stevens.edu
}
\input{header}
\begin{document}
\maketitle
\input{sections/abstract}
\input{sections/intro}

\input{sections/background}

\input{sections/preliminary}
\input{sections/RQ1}

\input{sections/RQ2}

\input{sections/related}

\input{sections/conclusion}

% \newpage

\section*{Limitations}

Although we conduct a comprehensive evaluation and analysis of early-exit behavior in modern \LLMs, several limitations remain.
First, our study focuses on a model’s \textbf{inherent} suitability for early-exit, considering methods that require little or no additional training or architectural changes.
In contrast, prior work~\cite{pan2024eetuning,xu2025specee} has shown that early-exit performance can be improved through finetuning or by introducing auxiliary prediction heads at intermediate layers, which can alter the model’s internal representations.
Second, due to the limited public availability of detailed pretraining recipes for most open-weight \LLMs, our analysis of training dynamics relies primarily on the released checkpoints of Pythia~\cite{biderman2023pythia}.
While these checkpoints provide useful insights, they offer only a partial view and make it difficult to draw general conclusions across model families.
In future work, we plan to incorporate early-exit–aware tuning techniques into our benchmark and conduct controlled small-scale pretraining experiments to better understand which training stages enhance or hinder a model’s suitability for early-exit.

% \section*{Acknowledgments} 

% \bibliography{custom,anthology-1,anthology-2}
\bibliography{custom}

% \bibliography{custom}

\input{sections/appendix}

\end{document}

%% file: header.tex
\usepackage{soul}
\usepackage{acronym}
\usepackage{xspace}
\usepackage{booktabs}
\usepackage{amsmath}
\usepackage{xcolor}
\usepackage{multirow}
\usepackage{algorithm}
\usepackage{algpseudocode} 

\soulregister\ac{1}
\soulregister\acs{1}
\soulregister\acl{1}
\soulregister\acf{1}

\sethlcolor{yellow}

\newcommand{\eg}{{\it e.g.}}

\newcommand{\vs}{\textit{vs.}\xspace}

\definecolor{ForestGreen}{RGB}{34,139,34}
\definecolor{RoyalBlue}{rgb}{0.25, 0.41, 0.88}

\input{_acronyms}

%% file: _acronyms.tex
\acrodef{ML}[ML]{Machine Learning}

\acrodef{NSF}[NSF]{National Science Foundation}

\acrodef{AI}[AI]{Artificial Intelligence}

\acrodef{FL}[FL]{Federated Learning}

\acrodef{CL}[CL]{Critical Learning}

\acrodef{AC}[AC]{Attacking-Critical}

\acrodef{CAGR}[CAGR]{compound annual growth rate}

\acrodef{CCT}[CCT]{Center for Computation and Technology}

\acrodef{SLO}[SLO]{service level objective}

\acrodefplural{SLOs}[SLO]{service level objectives}

\acrodef{RL}[RL]{reinforcement learning}

\acrodef{DRL}[DRL]{deep reinforcement learning}

\acrodef{VM}[VM]{virtual machine}

\acrodefplural{VM}[VMs]{virtual machines}

\acrodef{ITC}[ITC]{Innovation \& Technology Commercialization}

\acrodef{DAG}[DAG]{directed acyclic graph}

\acrodefplural{DAG}[DAGs]{directed acyclic graphs}

\acrodef{SFA}[SFA]{single point authentication}

\acrodef{HPC}[HPC]{high-performance computing}

\acrodef{SBIR}[SBIR]{Small Business Innovation Research}

\acrodef{IoT}[IoT]{Internet of Things}

\acrodef{DML}[DML]{distributed machine learning}

\acrodef{GNN}[GNN]{graph neural network}

\acrodefplural{GNN}[GNNs]{graph neural networks}

\acrodef{BSR}[BSR]{backdoor success rate}

\acrodef{BTA}[BTA]{backdoor task accuracy}

\acrodef{ATT}[ATT]{App Tracking Transparency}

\acrodef{DNN}[DNN]{deep neural network}

\acrodef{DNNs}[DNNs]{deep neural networks}

\acrodef{KL}[KL]{Kullback–Leibler}

\acrodef{IaaS}[IaaS]{Infrastructure-as-a-Service}

\acrodef{CaaS}[CaaS]{Container-as-a-Service}

\acrodef{TRPO}[TRPO]{Trust Region Policy Optimization}

\acrodef{CPO}[CPO]{Constrained Policy Optimization}

\acrodef{PPO}[PPO]{Proximal Policy Optimization}

\acrodef{TV}[TV]{Total Variation}

\acrodef{PAC}[PAC]{Probably Approximately
Correct}

\acrodef{ACI}[ACI]{Azure Container Instances}

\acrodef{NLP}[NLP]{Natural Language Processing}

\acrodef{GAE}[GAE]{Generalized Advantage Estimation}

\acrodef{IS}[IS]{Importance Sampling}

\acrodef{CV}[CV]{coefficient of variance}

\acrodefplural{CV}[CVs]{coefficient of variances}

\acrodef{LLM}[LLM]{Large Language Model}
\newcommand{\LLM}{\ac{LLM}\xspace}

\acrodef{GNS}[GNS]{gradient noise scale}

\acrodef{SOTA}[SOTA]{state-of-the-art}

\acrodef{SSP}[SSP]{Stale Synchronous Parallel}

\acrodef{CDF}[CDF]{cumulative distribution function}

\acrodef{PDF}[PDF]{probability density function}

\acrodef{RPC}[RPC]{remote procedure call}

\acrodef{MARL}[MARL]{multi-agent reinforcement learning}

\acrodef{SARL}[SARL]{single-agent reinforcement learning}

\acrodef{MDP}[MDP]{Markov Decision Process}

\acrodef{CTDE}[CTDE]{centralized training \& decentralized execution}

\acrodef{MAPD}[MAPD]{Multi-Agent Policy Distance}

\acrodef{IPPO}[IPPO]{Independent Proximal Policy Optimization}

\acrodef{MPE}[MPE]{Multi-Agent Particle Environment}

\acrodef{SMAC}[SMAC]{StarCraft Multi-Agent Challenge}

\acrodef{DDPG}[DDPG]{Deep Deterministic Policy Gradient}

\acrodef{DQN}[DQN]{Deep Q-Network}

\acrodef{MAPPO}[MAPPO]{Multi-Agent Proximal Policy Optimization}

\acrodef{MADDPG}[MADDPG]{Multi-Agent Deep Deterministic Policy Gradient}

\acrodef{IQL}[IQL]{Independent Q-Learning}

\acrodef{KDE}[KDE]{Kernel Density Estimation}

\acrodef{VDN}[VDN]{Value-Decomposition Networks}

\acrodef{SAC}[SAC]{Soft Actor-Critic}

\acrodef{QMIX}[QMIX]{Monotonic Value Function Factorisation}

\acrodef{IDDPG}[IDDPG]{Independent Deep Deterministic Policy Gradient}

\acrodef{RNN}[RNN]{recurrent neural network}

\acrodef{CNN}[CNN]{convolutional neural network}

\acrodef{MLP}[MLP]{multi-layer perception}

\acrodef{NN}[NN]{neural network}

\acrodef{PER}[PER]{Prioritized Experience Replay}

\acrodef{RLHF}[RLHF]{Reinforcement Learning from Human Feedback}

\acrodef{RLAIF}[RLAIF]{Reinforcement Learning from AI Feedback}

\acrodef{GRPO}[GRPO]{Group Relative Policy Optimization}

\acrodef{DPO}[DPO]{Direct Preference Optimization}

\acrodef{SFT}[SFT]{Supervised Fine-tuning}

\acrodef{FSDP}[FSDP]{Fully Sharded Data Parallel}

\acrodef{MoE}[MoE]{Mixture-of-Experts}
\newcommand{\MoE}{\ac{MoE}\xspace}

\acrodef{RM}[RM]{Reward Model}

\acrodef{OOM}[OOM]{out-of-memory}

\acrodef{KV}[KV]{key-value}

\acrodef{FLOPs}[FLOPs]{floating point operations}

\acrodef{TPOT}[TPOT]{time-per-output-token}
\newcommand{\TPOT}{\ac{TPOT}\xspace}

\acrodef{EWMA}[EWMA]{exponentially weighted moving average}

\acrodef{LoRA}[LoRA]{Low-Rank Adaption}

\acrodef{NCCL}[NCCL]{NVIDIA Collective Communications Library}

\acrodefplural{LLM}[LLMs]{Large Language Models}
\newcommand{\LLMs}{\acp{LLM}\xspace}

\acrodef{BERT}[BERT]{Bidirectional Encoder Representations from Transformers}

\acrodef{SSM}[SSM]{State Space Model}
\newcommand{\SSM}{\ac{SSM}\xspace}

\acrodefplural{SSM}[SSMs]{State Space Models}
\newcommand{\SSMs}{\acp{SSM}\xspace}

\acrodef{EAS}[EAS]{Early-exit adaptability score}
\newcommand{\EAS}{\ac{EAS}\xspace}

%% file: sections/abstract.tex
\begin{abstract}
In Large Language Model (LLM) inference, early-exit refers to stopping computation at an intermediate layer once the prediction is sufficiently confident, thereby reducing latency and cost. However, recent LLMs adopt improved pretraining recipes and architectures that reduce layer redundancy, potentially limiting early-exit opportunities. We re-evaluate layer-wise early-exit in modern LLMs and analyze how intermediate representations evolve during training. We introduce a metric to quantify a model’s intrinsic suitability for early-exit and propose a benchmark for researchers to explore the potential early-exit benefits on different models and workloads. 
Our results show a diminishing trend in early-exit effectiveness across newer model generations. We further find that dense transformers generally offer greater early-exit potential than Mixture-of-Experts and State Space Models. In addition, larger models, particularly those with more than 20 billion parameters, and base pretrained models without specialized tuning tend to exhibit higher early-exit potential.
\end{abstract}

%% file: sections/intro.tex
\section{Introduction}
\label{sec:intro}

\begin{figure}[t]
  \centering
  \includegraphics[width=.95\linewidth]{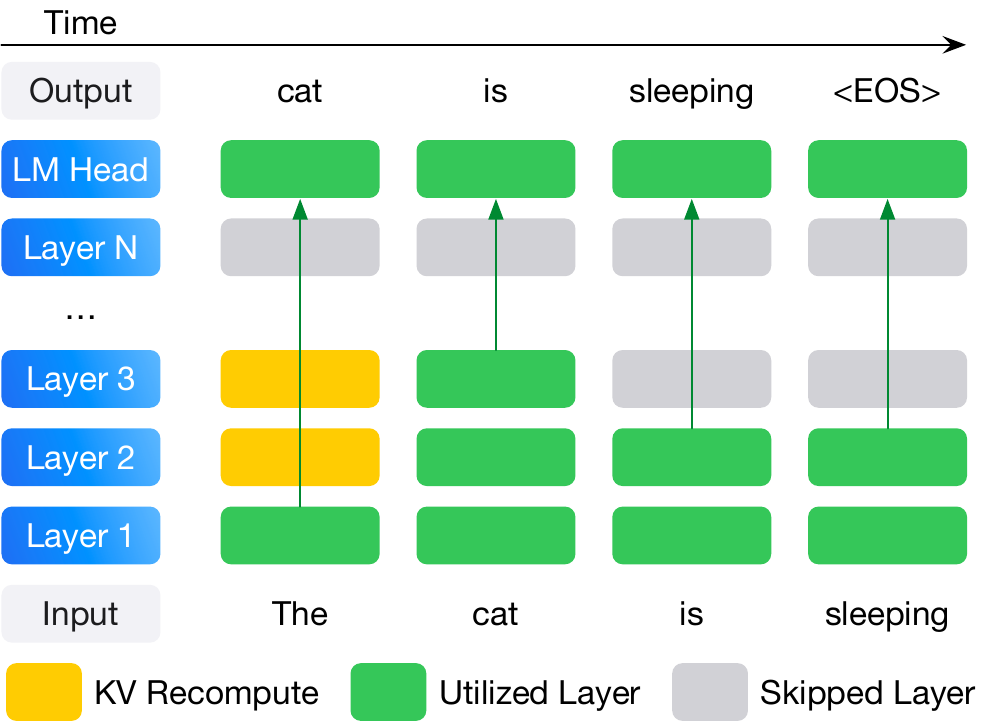}
  \vspace{-0.1in}
  \caption{
Layer-wise early-exit decoding in LLMs. 
% (a) The dataflow of layer-wise early-exit mechanisms. (b) T. (c) The research questions this paper is trying to address.
  } 
  \vspace{-0.2in}
  \label{fig:intro_ee}
\end{figure}

In \LLM inference, early exit refers to terminating computation at an intermediate layer when the model has achieved sufficient confidence, which improves efficiency by lowering latency and computational cost.
Early-exit has been widely studied as a mechanism to reduce inference cost by allowing models to terminate computation at intermediate layers~\cite{chen2023eellm,liu-etal-2024-speculative-decoding,pan2024eetuning,elhoushi-etal-2024-layerskip,xu2025specee}, as shown in Fig.~\ref{fig:intro_ee}. This idea has been successfully applied to traditional machine learning models and earlier generations of \LLMs~\cite{touvron2023llama2openfoundation}, where many layers are redundant and intermediate representations often contain sufficient information to produce accurate predictions. 
Prior work shows that early-exit can significantly reduce latency and computation under acceptable accuracy loss.

% \hnote{can we also have stats of papers related to ``early-exit'' to show the trend of research?}
% \hnote{For the storyline, we need to identify what is the exact gap here: 1) why do \textbf{so many} studies work on early exit? 2) how the opportunity for early exit is diminishing? 3) discussion on future trends and opportunity (how this proposed score/metric will help?)}\ricky{This metric can help people estimate the potential benefits of applying early-exit to a specific model, before they actually spend time and resources designing and implementing it?}
% %
% \hanfei{I suggest moving Figure 2a up right on the first page and then followed by Figure 1.}

However, recent \LLMs~\cite{meta_llama4_2025,yang2025qwen3technicalreport} differ substantially from earlier models in both architecture and training methodology. Modern models adopt improved pretraining recipes, larger and more diverse datasets, and architectural changes such as stronger normalization, deeper networks, and \MoE designs, which can shift intermediate computation toward later layers and change the relationship between middle layers and the final layer. As a result, assumptions that motivated earlier early-exit methods---namely that certain intermediate layers are highly redundant and can reliably approximate final-layer outputs---may no longer hold.
Moreover, most existing early-exit methods are tightly coupled to specific models or workloads, and typically require non-trivial design and tuning effort for each model or task~\cite{elhoushi-etal-2024-layerskip,pan2024eetuning,chen2023eellm,xu2025specee}. As a result, there is currently no general mechanism to estimate or quantify the potential benefits of early-exit before committing to a particular design.
% Most existing early-exit methods implicitly assume that intermediate layers can generate outputs that are semantically close to those produced by the final layer. If this assumption fails, early-exit either leads to severe accuracy degradation or requires additional fine-tuning and retraining, which limits its practicality. Moreover, production-level \LLM serving imposes strict constraints on batching and throughput, further complicating the deployment of early-exit mechanisms.

\begin{figure}[t]
  \centering
  \includegraphics[width=.98\linewidth]{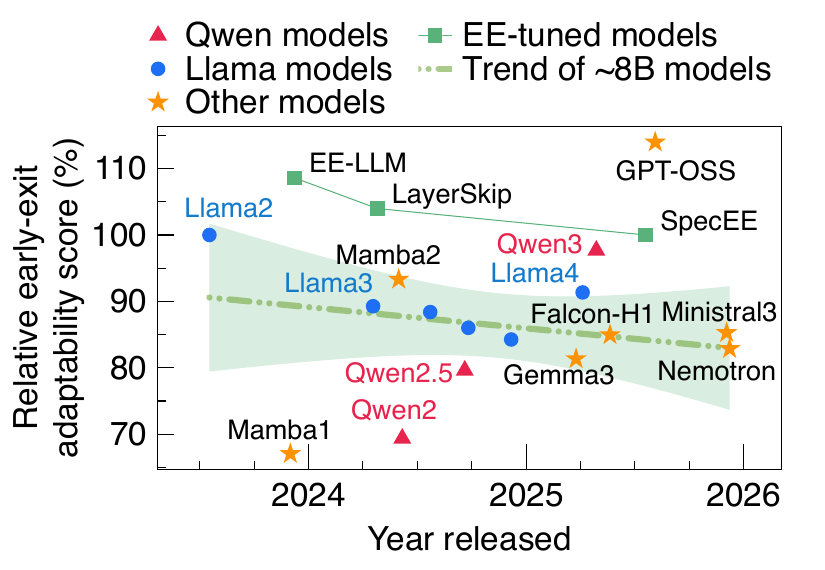}
  \vspace{-0.15in}
  \caption{
The trend of relative early-exit scores (\S\ref{subsec:metrics}) in recent \LLMs and models specifically tuned for early-exit% ~\cite{touvron2023llama2openfoundation,gu2024mambalineartimesequencemodeling,dao2024transformersssmsgeneralizedmodels,grattafiori2024llama3herdmodels,yang2024qwen2technicalreport,meta_llama4_2025,yang2025qwen3technicalreport,openai2025gptoss120bgptoss20bmodel,Mistral3,gemma_2025,falconh1,Nemotron_Cascade}, 
, compared to \texttt{Llama2-7B}. 
% The early-exit score metric is defined in \S\ref{subsec:metrics}. 
% The green dashed line and shadow area show the early-exit score trend of base models with approximately 8 billion parameters. 
We explain the model selection details in Appendix \ref{appendix:evaluate_ee_score}.
% \hnote{first show what is early exit as Fig 2(a), and then show this trend} \todo{Show an aggregated result?}
  } 
  \vspace{-0.2in}
  \label{fig:intro}
\end{figure}

In this paper, we revisit layer-wise early-exit decoding in the context of recent \LLMs. We systematically evaluate whether modern \LLMs still exhibit exploitable early-exit opportunities for the decoding stage, and analyze factors that affect early-exit effectiveness. 
Our main contribution includes:
% \begin{itemize}[noitemsep,topsep=0pt,parsep=0pt,partopsep=0pt]
\begin{itemize}[leftmargin=*, nosep]
    \item We introduce a new metric, the \textit{early-exit adaptability score} (\S\ref{subsec:metrics}), and a benchmark framework with oracle early-exit evaluation (\S\ref{subsec:sim-vs-acc}). Together, they quantify a model’s intrinsic suitability for early exit and estimate its upper-bound acceleration potential. We plan to open-source the benchmark after the paper is accepted.

    \item \textbf{Observation 1:} We report the early-exit scores across multiple model generations and observe a decreasing trend as \LLMs evolve (Fig.~\ref{fig:intro}), indicating that newer models exhibit reduced layer redundancy and are less amenable to early-exit.
    
    \item \textbf{Observation 2:} We further analyze the main factors that could shape the early-exit phenomenon. We find that the early-exit behavior is mainly shaped by four factors: (1) larger models generally exhibit higher early-exit suitability; (2) dense transformers are more amenable to early-exit than \MoE and \SSMs; (3) continued pretraining and post-training tuning tend to reduce early-exit suitability; and (4) early-exit patterns are largely model-specific and only weakly influenced by the assigned workload.
\end{itemize}

%% file: sections/background.tex
\section{Background \& Motivation}

\begin{figure}[t]
  \centering
  \includegraphics[width=.98\linewidth]{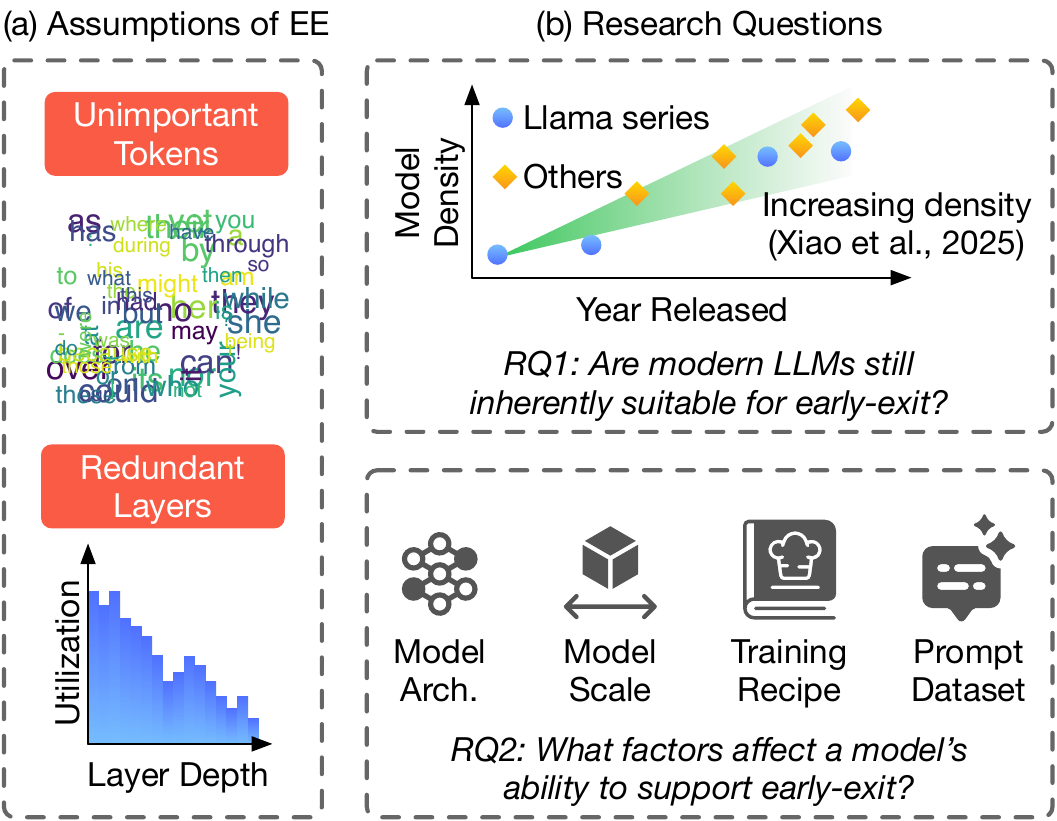}
  \vspace{-0.1in}
  \caption{
Workflow illustration of the paper. 
% (a) The dataflow of layer-wise early-exit mechanisms. (b) T. (c) The research questions this paper is trying to address.
  } 
  \vspace{-0.2in}
  \label{fig:overview}
\end{figure}

\subsection{Early-Exit in Large Language Models}

Early-exit methods aim to reduce inference latency by terminating computation before reaching the final layer when intermediate representations are deemed sufficient, as shown in Fig.~\ref{fig:overview}(a). In the context of \LLMs, this can potentially reduce the \TPOT latency, which is critical for large-scale serving deployment.
% 
% \hl{Prior to early-exit in \LLMs}, similar techniques have been widely applied in traditional \ML methods~\cite{graves2017adaptivecomputationtimerecurrent,teerapittayanon2017branchynetfastinferenceearly,zhou2020bertlosespatiencefast}. \hnote{do we really need to mention this?}
% 
Recent early-exit methods for \LLMs, such as SpecEE~\cite{xu2025specee} and EE-LLM~\cite{chen2023eellm,pan2024eetuning}, introduce auxiliary exit heads or confidence-based criteria to decide when to terminate inference early. These approaches demonstrate promising results on earlier \LLM generations, such as Llama2~\cite{touvron2023llama2openfoundation}, GPT-2~\cite{radford2019gpt2}, and Vicuna~\cite{zheng2023vicuna}.

\subsection{Limitations of Existing Early-Exit Assumptions}

The benefits of early-exit methods come from finding the sweet spot when doing a trade-off between accuracy and inference latency. The core goal is to find out the earliest position to stop the inference without harming the generation quality. 
Most early-exit methods rely on two key assumptions: 

\textbf{Many tokens contribute little to the final output.}
These tokens are often referred to as \emph{easy tokens}~\cite{schuster2022confidentadaptivelanguagemodeling,bae-etal-2023-fast}, meaning that their predicted logits stabilize early and can be generated accurately using intermediate layers rather than the full model.  
However, the overall benefit of early-exit depends on how many such easy tokens appear during generation. If only a small fraction of tokens can exit early, the resulting speedup will be limited. Moreover, recent advances in pretraining encourage more uniform information flow across layers~\cite{yang2024qwen2technicalreport,grattafiori2024llama3herdmodels,openai2025gptoss120bgptoss20bmodel}, which may delay token stabilization and further reduce the opportunities for early exit.

\textbf{The model is effectively sparse at inference time,} meaning that many layers are redundant for generating a single specific token.
% , which has a limited influence on the complete output.
% 
Under this assumption, intermediate layers can be directly mapped to the language model head to produce meaningful output tokens.
While this assumption has been empirically validated for earlier \LLMs~\cite{touvron2023llama2openfoundation}, recent models are explicitly designed to reduce such redundancy with increasing density~\cite{xiao2025densing}.
% 
% \MoE architectures are also proposed to selectively activate parameters to improve efficiency~\cite{yang2025qwen3technicalreport,meta_llama4_2025}, with dense computation inside each expert. 
% 
As a result, intermediate layers may no longer produce logits that align well with the final-layer output.
% , leading to accuracy degradation under early-exit. 
% 

For early-exit to preserve output quality, the results produced at an exit layer must induce token distributions similar to those of the final layer, which is further validated in \S\ref{subsec:sim-vs-acc}. 
If intermediate layers generate poorly calibrated or semantically inconsistent distributions that diverge across layers, early-exit results in substantial accuracy degradation.
Alternatively, tokens may have to exit at very late layers to preserve output quality, which largely eliminates the latency benefits of early termination (\S\ref{subsec:upper_bound}).
In such cases, additional fine-tuning or retraining is required to adapt the model to an early-exit setting~\cite{pan2024eetuning,elhoushi-etal-2024-layerskip}, which substantially amplifies system complexity and training cost, and alters the model behavior to maintain acceptable generation quality.

% \hnote{We need experiments to show the assumptions' limitations}
% \ricky{The assumptions do not hold for newer LLMs, which will be shown in later sections where we evaluate the accuracy under different similarity threshold}
% \hanfei{If not showing detailed results, maybe map each point and refer to the later specific sections?}

% \subsection{Challenges for Production-Level Serving}

% Beyond accuracy concerns, early-exit introduces system-level challenges. In production serving, tokens are typically batched to maximize throughput~\cite{}. Layer-wise early-exit causes tokens within the same batch to terminate at different layers, preventing efficient batching and requiring dynamic execution paths. This leads to underutilized hardware and limits practical latency gains.

% Additionally, existing early-exit methods often exhibit an unfavorable accuracy-latency trade-off when applied to modern \LLMs under realistic serving conditions. These limitations motivate a systematic re-evaluation of early-exit in recent models and an investigation into the factors that determine whether early-exit remains viable.

\subsection{Research Questions}

To further understand the early-exit phenomena in a systematic way, the research questions listed below have to be answered:

\textbf{RQ1: Are modern \LLMs still inherently suitable for layer-wise early-exit?}
Recent decoder-only \LLMs differ substantially from earlier models in architecture and training, which can alter how predictive signals evolve across layers during decoding.
In this work, we evaluate whether intermediate layers produce outputs consistent with the final layer by measuring the similarity of output logits, top-$K$ token predictions, and hidden representations across layers for varying \LLMs.

\textbf{RQ2: What factors affect a model’s ability to support early-exit?}
Regardless of whether modern \LLMs remain suitable for early-exit, it is important to understand the underlying factors that drive this behavior. We analyze how model architecture, scale, training schemes, and generation characteristics influence the effectiveness of early-exit, to identify conditions under which model efficiency can be improved by early-exit mechanism.

% \textbf{RQ3: What are the possible}
% 

%% file: sections/preliminary.tex
\section{Methodology}

\subsection{Datasets}
% \hnote{Should 3.1 and 3.2 be placed to Evaluation? I am not familiar with this writing structure. Do we have any references (existing papers) for such a structure?}

% 
To study how workload characteristics affect early-exit behavior, we select datasets that vary in output length and task type. Longer outputs may offer greater latency savings from early exit, but accuracy can degrade as token-level errors accumulate during decoding~\cite{arora-etal-2022-exposure,gan2025rethinking}. Different task types may also rely on model depth in different ways, leading to distinct early-exit behaviors. We evaluate GPQA~\cite{rein2023gpqagraduatelevelgoogleproofqa}, GSM8K~\cite{cobbe2021gsm8k}, HumanEval~\cite{chen2021humaneval}, and MMLU~\cite{hendrycks2021mmlu}, which cover scientific reasoning, mathematical problem solving, code generation, and short-form knowledge evaluation. These datasets span both long-form reasoning tasks and short, decision-focused tasks with concise outputs. For each dataset, we use a fixed subset of 100 prompts selected with the same random seed across all models to ensure fair comparisons while keeping the experimental matrix tractable.

% We additionally verify robustness by repeating key experiments with larger subsets for a representative model, observing consistent trends.

\subsection{Models}

To study how model scale, architecture, and training recipe affect early-exit behavior, we evaluate a diverse set of modern language models with open-sourced weights. Our selection includes (1) Meta’s Llama2~\cite{touvron2023llama2openfoundation}, Llama3~\cite{grattafiori2024llama3herdmodels}, and Llama4~\cite{meta_llama4_2025} families, (2) Alibaba’s Qwen2~\cite{yang2024qwen2technicalreport} and Qwen3~\cite{yang2025qwen3technicalreport} models, (3) OpenAI’s OSS model~\cite{openai2025gptoss120bgptoss20bmodel}, and (4) Mamba1~\cite{gu2024mambalineartimesequencemodeling} and Mamba2~\cite{dao2024transformersssmsgeneralizedmodels} models.
These models span a wide range of scales and architectural designs. We summarize the detailed model comparison in \ref{appendix:model_comparison}.
% 
% As for training recipes, most providers do not fully disclose detailed training configurations. Therefore, we report only the publicly available information released through technical reports, model cards, and official documentation.

\subsection{Metrics}
\label{subsec:metrics}

To measure the model's potential ability to adapt the early-exit mechanism directly in an intuitive manner, a new metric must be defined, considering both the model acceleration and the accuracy loss.

\textbf{Skip ratio.}
To estimate the acceleration, we define the skip ratio at the exit layer $\ell$ as
$
w_{\ell}=\frac{L-\ell}{L},
$
where $L$ is the total number of layers. A larger $w_{\ell}$ indicates more skipped layers and higher potential speedup~\cite{xu2025specee,chen2023eellm}.

\textbf{Layer-to-final similarity.}
End-to-end accuracy under early-exit is only observable at discrete exit thresholds. To obtain a continuous proxy for output quality, we measure the cosine similarity between the hidden state, which can also be replaced with the output logit or top-$K$ candidates, at exit layer $\ell$ and the final layer $L$:
$
S_\ell = \frac{\mathbf{h}_\ell \cdot \mathbf{h}_L}{\|\mathbf{h}_\ell\| \,\|\mathbf{h}_L\|}.
$
Higher $S_\ell$ indicates that the exit-layer representation is closer to the final-layer representation, similar to prior layer-wise analysis work~\cite{men-etal-2025-shortgpt,csordás2025languagemodelsusedepth}.
We further validate that the similarities can reflect the model's actual performance under early-exit settings in \S\ref{subsec:upper_bound}.

\textbf{\EAS.}
Early-exit introduces a classic accuracy-efficiency trade-off, which is a standard multi-objective optimization setting. An intuitive way to obtain a single score is to use a scalarization that rewards both objectives while penalizing imbalance.
We define an adaptability score $A_\ell$ for exiting at layer $\ell$ as a weighted geometric mean that balances efficiency and accuracy. The skip ratio represents the potential acceleration from early termination, while the similarity term captures how closely the exit-layer output matches the full-depth model and thus reflects the expected accuracy under early-exit.
To ensure compatibility with the skip ratio and to bound the adaptability score within $[0,1]$, we first map the layer-to-final similarity to the unit interval using a monotonic mapping function $f(\cdot)$:
\begin{equation*}
\tilde{S}_\ell = f(S_\ell), \;\;\; \tilde{S}_\ell \in [0,1].
\end{equation*}
The adaptability score is then defined as
\begin{equation*}
A_\ell = \tilde{S}_\ell^{\alpha}\cdot w_\ell^{1-\alpha}, \;\;\; \alpha\in[0,1].
\end{equation*}

In this work, we use a linear scaling 
% $f(S_\ell) = (S_\ell + 1)/2$ 
to map cosine similarity to $[0,1]$ for simplicity.
% This choice preserves the relative ordering of similarity values, 
% % avoids introducing additional hyperparameters, 
% and provides a simple and interpretable normalization.
% Other monotonic mappings (e.g., clipping or exponential transforms) can be used without affecting the general formulation.
% 
To summarize a model over all candidate exit layers, we report the average early-exit adaptability score as
\begin{equation*}
\mathrm{EAS} = \sum_{\ell\in\{1,\dots,L-1\}} \frac{A_\ell}{L-1}.
\end{equation*}

\subsection{Evaluation Setup}

We implement our benchmarking framework on top of OpenCompass~\cite{2023opencompass}, which is an open-source \LLM benchmarking system with tuned instructions and unified evaluation methods for different datasets.
Following prior evaluation methodology~\cite{chen-etal-2025-clasp,elhoushi-etal-2024-layerskip}, we set the model temperature to zero to ensure deterministic results for consistent benchmarking.
The performances of the downstream tasks are evaluated in a zero-shot manner.
The default maximal output limit is set to 1024 tokens.

%% file: sections/RQ1.tex
\section{RQ1: Evaluating Modern \LLMs' Adaptability to Early-Exit}
\label{sec:rq1}

In this section, we evaluate whether modern \LLMs can inherently support traditional early-exit mechanisms, and analyze the resulting trade-offs between inference acceleration and accuracy across different model families and generations.
Our evaluation focuses on three key aspects.
First, we analyze \textbf{(1) layer-to-final similarity} to understand how predictive information is distributed across layers and to quantify the degree of layer redundancy in different generations of \LLMs during token generation.
Second, we study the \textbf{(2) relationship between layer-to-final similarity and early-exit accuracy} to examine whether similarity measurements reliably reflect actual early-exit performance.
Finally, we estimate \textbf{(3) the upper bound of early-exit benefits}, defined as the maximum achievable layer skip ratio under the constraint of maintaining the original generation behavior with acceptable accuracy loss.
Together, these analyses allow us to assess whether layer redundancy in modern \LLMs can be effectively exploited for acceleration and how it translates to practical early-exit performance.

% \subsection{Experimental Setup}

\subsection{Layer-to-Final Similarity Analysis}

\begin{figure*}[t]
  \centering
  \includegraphics[width=.97\linewidth]{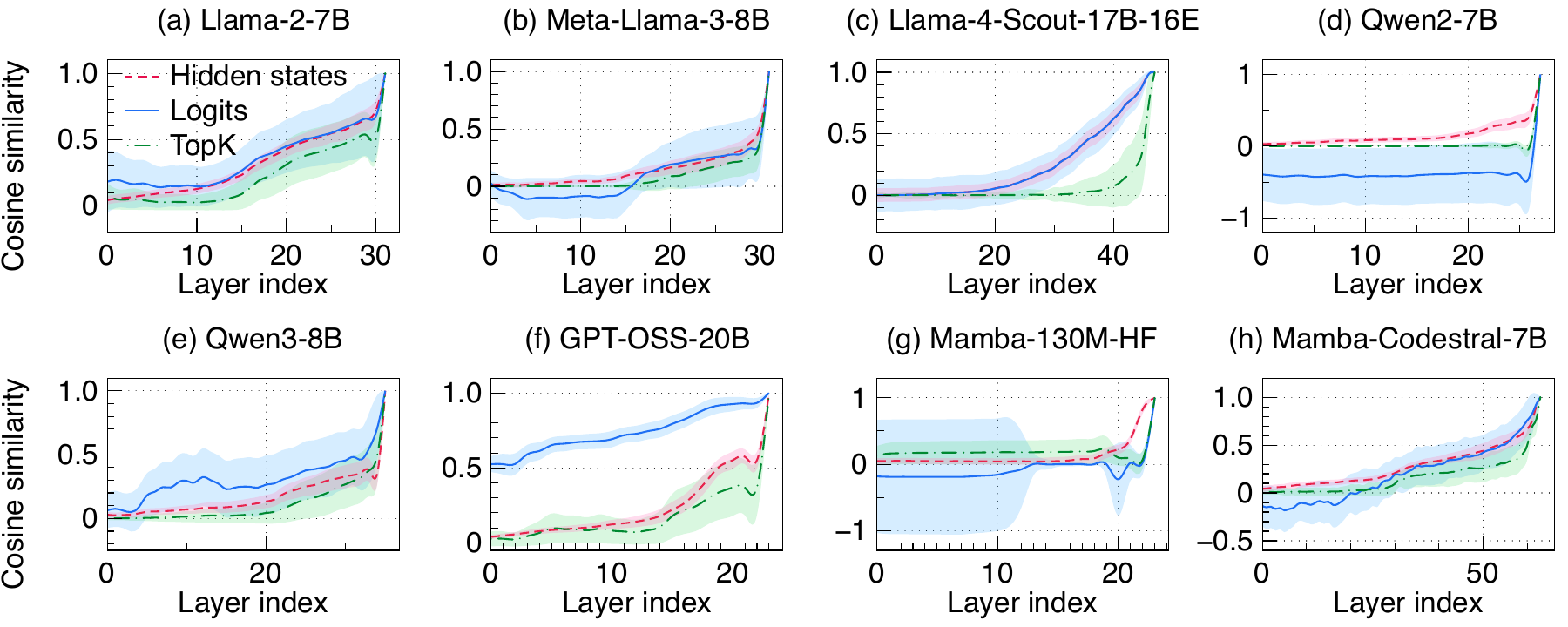}
  \vspace{-0.15in}
  \caption{
The layer-to-final similarity results of eight different models
% ~\cite{touvron2023llama2openfoundation,grattafiori2024llama3herdmodels,meta_llama4_2025,yang2024qwen2technicalreport,yang2025qwen3technicalreport,openai2025gptoss120bgptoss20bmodel,gu2024mambalineartimesequencemodeling,dao2024transformersssmsgeneralizedmodels}
aggregated across four datasets.
% ~\cite{cobbe2021gsm8k,rein2023gpqagraduatelevelgoogleproofqa,hendrycks2021mmlu,chen2021humaneval}
% The shadow area demonstrates the standard deviation.
  } 
  \vspace{-0.2in}
  \label{fig:eval_R1_similarity}
\end{figure*}

\subsubsection{Experimental Setup}

To preserve the original generation behavior of a model that uses full-depth decoding, the information produced at an exit layer must be highly consistent with that of the final layer.
Accordingly, we measure the cosine similarity between exit layers and the final layer using three signals: 

\textbf{Hidden state (semantic)} similarity measures how close the intermediate-layer representations are to the final-layer representation, indicating whether the model has formed a final-like semantic state for next-token prediction.
Higher hidden state similarity suggests that later layers mainly refine existing information, which is a necessary but not sufficient condition for early-exit.

\textbf{Output logits (probability)} measure how closely the next-token probability distribution produced by an intermediate layer matches that of the final layer. For intermediate layers, logits are obtained by applying the language model head to their hidden states, projecting them into the vocabulary space. This directly captures whether an intermediate layer would make the same next-token prediction as the final layer, which is required to preserve decoding decisions under early-exit inference.
% \hanfei{Do we need to describe a bit more to distinguish between hidden state and output logits? This probably needs to mention LM head}
% \hanfei{Do we need to show how and where we get those intermediate states in Figure 2a?}

\textbf{Top-$K$ token predictions (candidate-set)} measures the overlap between the candidate sets predicted at an intermediate layer and the final layer, capturing candidate-set stability rather than exact probabilities.
High top-$K$ similarity suggests that early-exit may still be feasible under relaxed decoding settings where minor probability shifts are tolerated.
We set $K=10$ to provide a robust stability measure while avoiding minor perturbations.

\subsubsection{Key Observations}

Figure~\ref{fig:eval_R1_similarity} presents the layer-to-final similarity results and highlights the degree of layer redundancy across model generations.
In general, higher similarity emerging at earlier layers indicates greater intrinsic potential for integrating early-exit mechanisms without significantly affecting output quality.

\textbf{Late and gradual similarity growth.}
In Fig.~\ref{fig:eval_R1_similarity}, most models exhibit low layer-to-final similarity in early layers, followed by a gradual increase toward the final layers.
This indicates that representations and decisions are progressively refined, with final-like behavior emerging only near the end of the network.
Such step-by-step refinement is consistent with prior layer-wise analyses~\cite{yan-2025-addition,yu-etal-2025-back,skean2025layer}.
Compared to newer models, older dense models such as \texttt{Llama2-7B} show a smoother and earlier rise in similarity, while newer models (e.g., \texttt{Llama-3-8B} and \texttt{Qwen} variants) remain below moderate similarity levels for most layers and increase sharply only near the end.
This suggests reduced representation-level redundancy in newer models and a narrower safe window for adopting early-exit techniques.

\textbf{Stable semantic flow across inputs.}
Across almost all models, the standard deviation of hidden-state similarity is relatively small, indicating that the layer-wise semantic evolution is largely a model-level property rather than being strongly affected by individual tokens or sequences.
% The small variance across tokens indicates that the depth-wise semantic evolution is stable in aggregate and largely determined by the model rather than individual inputs.
This stability implies that the depth-wise processing pattern is consistent across inputs, making hidden-state trends reliable for comparative analysis across models. \citet{csordás2025languagemodelsusedepth} report similar layer-wise patterns on Llama3 and Qwen3 models, and our results extend these observations to a broader set of architectures and generations, including Qwen2, Llama4, GPT-OSS, and Mamba families.

\textbf{Early logit alignment in the OSS model.}
In \texttt{GPT-OSS-20B}, output logit similarity remains high and stable across layers, while top-$K$ similarity exhibits larger fluctuations.
This indicates that the overall next-token probability distribution is calibrated early, whereas small probability differences among closely competing tokens lead to frequent rank changes near the top-$K$ boundary.
Such behavior suggests that the OSS model refines predictions in a smooth and incremental manner rather than applying aggressive late-stage corrections.
In contrast, many other models tend to perform stronger probability recalibration in deeper layers, causing logit and top-$K$ similarities to increase more synchronously.
Overall, the early stabilization of logit distributions implies that the OSS model is more amenable to early-exit mechanisms without significant accuracy degradation.

\begin{table*}[t]
\centering
\small
\caption{Oracle early-exit performance under a maximum skip ratio, with accuracy loss within to 5\%.
We report the full-depth accuracy (Full Acc.$\uparrow$), early-exit accuracy (Acc.$\uparrow$), skip ratio (Skip$\uparrow$), and the mean early-exit score ($\uparrow$).}
\vspace{-0.1in}
\label{tab:oracle_performance}
\setlength{\tabcolsep}{7pt}
\renewcommand{\arraystretch}{1.15}
\begin{tabular}{c | cccc | cccc | c}
\toprule
\multirow{2}{*}{\textbf{Model}} 
& \multicolumn{4}{c|}{\textbf{MMLU}} 
& \multicolumn{4}{c|}{\textbf{GSM8K}} 
& \multirow{2}{*}{\textbf{\EAS}} \\
\cmidrule(lr){2-5}\cmidrule(lr){6-9}
& \textbf{$\delta$} & \textbf{Full Acc.} & \textbf{Acc.} & \textbf{Skip (\%)} 
& \textbf{$\delta$} & \textbf{Full Acc.} & \textbf{Acc.} & \textbf{Skip (\%)} 
& \\
\midrule

Llama2-7B
& 0.70 & 0.41 & 0.38 & 4.69
& 0.80 & 0.21 & 0.17 & 2.13
& 0.52 \\

Llama3-8B
& 0.91 & 0.54 & 0.52 & 0.18
& 0.90 & 0.37 & 0.33 & 0.04
& 0.46 \\

Llama4-Scout
& 0.84 & 0.76 & 0.73 & 0.32
& 0.87 & 0.85 & 0.82 & 0.18
& 0.47 \\

Qwen2-7B
& 1.00 & 0.24 & 0.24 & 0.00
& 1.00 & 0.53 & 0.53 & 0.00
& 0.36 \\

Qwen3-8B
& 0.95 & 0.41 & 0.39 & 0.10
& 0.86 & 0.54 & 0.51 & 0.62
& 0.51 \\

OSS-20B
& 0.96 & 0.74 & 0.69 & 7.26
& 0.93 & 0.61 & 0.57 & 4.48
& 0.59 \\

\bottomrule
\end{tabular}

\vspace{-0.15in}
\end{table*}

\textbf{Atypical logit behavior in specific models.}
Two models show notably different patterns from the rest.
For \texttt{Qwen2-7B}, logit similarity remains low or negative for most layers while top-$K$ similarity increases later, indicating that although candidate tokens may overlap, their relative probabilities differ substantially from the final distribution.
This suggests strong late-stage recalibration of logits, which limits safe early-exit despite partial candidate stability.
For the small-capacity \texttt{Mamba-130M} model, logit similarity exhibits large fluctuations and high variance across layers, likely due to limited model capacity and unstable intermediate representations.
As a result, similarity-based early-exit signals are less reliable for such small models.

\subsection{Layer-to-Final Similarity \vs Accuracy}
\label{subsec:sim-vs-acc}

\begin{figure}[t]
  \centering
  \includegraphics[width=.95\linewidth]{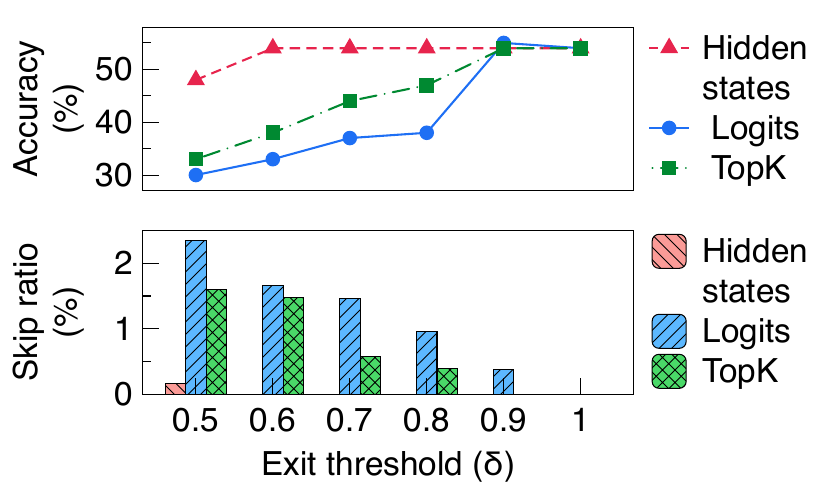}
  \vspace{-0.15in}
  \caption{
The accuracy and skip ratio of different exit strategies and thresholds.
  } 
  \vspace{-0.2in}
  \label{fig:eval_R1_accuracy_trend}
\end{figure}

To evaluate whether layer-to-final similarity can reliably predict end-to-end accuracy under early-exit, and to assess the impact of different similarity signals on early-exit suitability, we conduct \textit{oracle} experiments in which similarity information is assumed to be available when selecting the exit layer.
In this setting, the model exits at a layer only if the current similarity exceeds a predefined threshold $\delta$.
Figure~\ref{fig:eval_R1_accuracy_trend} shows the resulting accuracies and skip ratios for different similarity-based oracle early-exit strategies evaluated on \texttt{Qwen3-8B} using the GSM8k~\cite{cobbe2021gsm8k} dataset.

\textbf{Logit similarity affects the early-exit performance the most.}
The results show clear differences in how similarity signals trade accuracy for acceleration under early-exit.
% Hidden-state–based criteria maintain stable accuracy across thresholds but yield almost no skip ratio, indicating that semantic similarity alone is insufficient to enable effective early exits.
Logit-based early-exit exhibits a clear accuracy--efficiency trade-off: low thresholds achieve high skip ratios with severe accuracy loss, while higher thresholds recover accuracy at the cost of reduced acceleration.
In contrast, hidden-state-based and top-$K$–based criteria preserve accuracy across thresholds but provide only marginal skip benefits, suggesting that candidate tokens and semantic states stabilize late, consistent with Fig.~\ref{fig:eval_R1_similarity}.
Overall, logit similarity is the most sensitive signal for early-exit control. Thus, we select it as the core component of the \EAS (\S\ref{subsec:metrics}).

\subsection{Upper Bound Exploration}
\label{subsec:upper_bound}

The similarity analysis provides an internal, model-centric view for understanding how different generations of \LLMs differ in their layer-wise behavior.
However, similarity between an exit layer and the final layer does not directly translate to end-to-end task accuracy when early-exit is applied.
To further examine the upper bound of conventional early-exit methods, we conduct more oracle experiments on the MMLU~\cite{hendrycks2021mmlu} and GSM8k~\cite{cobbe2021gsm8k} datasets, using the logit-similarity-based exit criteria, in which the model can make more informed exit decisions than commonly used confidence-based criteria~\cite{schuster2022confidentadaptivelanguagemodeling,chen2023eellm,zhou2020bertlosespatiencefast}, thereby revealing the maximum achievable early-exit potential of selected \LLMs.
We report the oracle early-exit results in Table~\ref{tab:oracle_performance}, showing that the \EAS is broadly consistent with the achievable early-exit benefits.
The exit thresholds $\delta$ are selected using a simple linear search to maximize the skip ratio while keeping the accuracy loss within 5\%.
The results show that directly applying early-exit mechanisms to base models, even when layer-to-final similarity information is assumed to be known in advance, fails to achieve a balance between accuracy and acceleration in \LLMs.

% \todo{Add a paragraph of analysis}

%% file: sections/RQ2.tex
\section{RQ2: Evaluating Factors that Affect LLMs' Early-Exit Opportunity}
\label{sec:rq2}

To fundamentally understand what causes the early-exit behaviors of \LLMs shown in \S\ref{sec:rq1}, we further evaluate the models using the \EAS metric and analyze the correlation between model or inference settings and the early-exit behavior.

% \subsection{Model}

\subsection{Model Scale}

\begin{figure}[t]
  \centering
  \includegraphics[width=.95\linewidth]{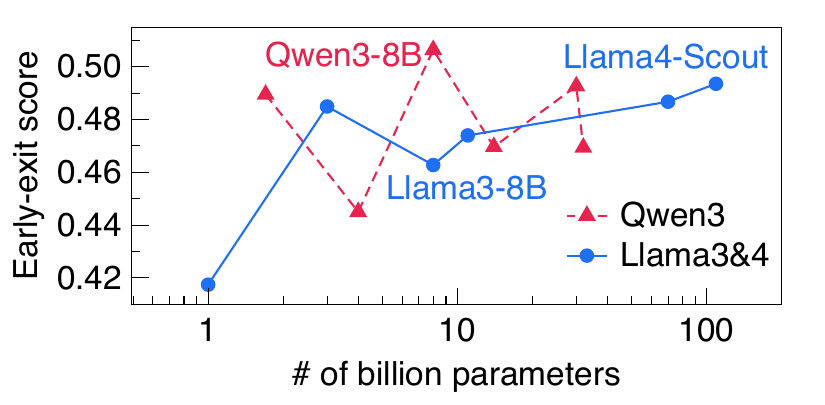}
  \vspace{-0.15in}
  \caption{
The \EAS increases with model scale.
  } 
  \vspace{-0.1in}
  \label{fig:evcal_R2_scale}
\end{figure}

The first aspect of the model that can affect the model's ability to early-exit is the scale (\eg, \# of parameters and \# of layers).
In fact, many \LLMs' smaller flavors are distilled from the largest one, and are usually denser than larger models, which can potentially reduce the availability for early-exit.
To validate if the previous observation in \S\ref{sec:rq1} still exist in larger models, and to figure out if the early-exit patterns will be affected by the scale of the \LLM, we further run experiments using models with varying sizes from the Qwen3~\cite{yang2025qwen3technicalreport}, Llama3~\cite{grattafiori2024llama3herdmodels}, and Llama4~\cite{meta_llama4_2025} families.
Fig.~\ref{fig:evcal_R2_scale} shows that, in most cases, \textit{a model’s early-exit suitability increases with model scale}, as larger models tend to exhibit greater layer-wise redundancy due to the addition of more layers and parameters.

\subsection{Model Architecture}

\begin{figure}[t]
  \centering
  \includegraphics[width=.85\linewidth]{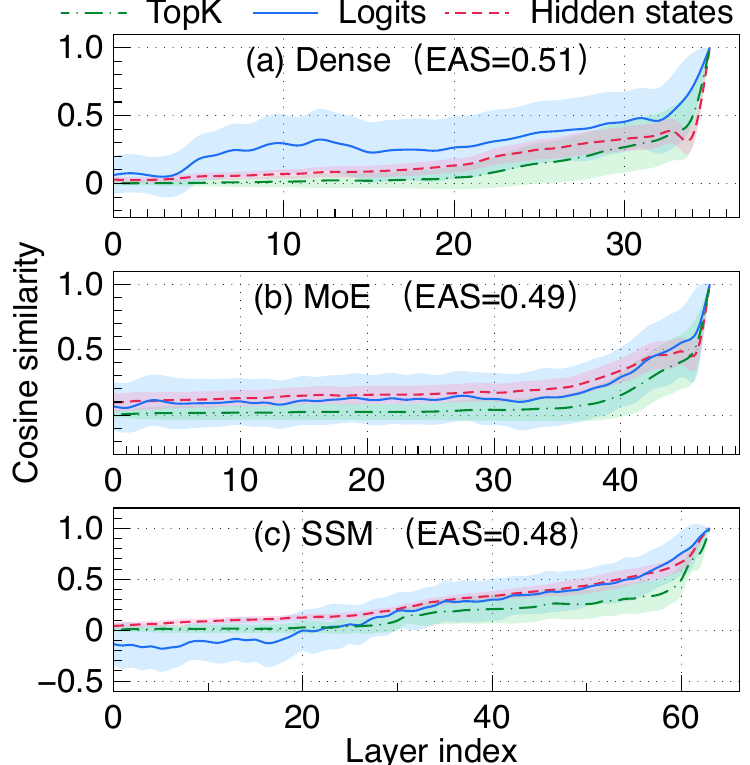}
  \vspace{-0.1in}
  \caption{
The layer-to-final similarity and the mean \EAS of three different \LLM architectures: 
(a) dense (\texttt{Qwen3-8B}),
% ~\cite{yang2025qwen3technicalreport}, 
(b) \MoE (\texttt{Qwen3-30B-A3B}),
% ~\cite{yang2025qwen3technicalreport}, 
and (c) \SSM (\texttt{Mamba-Codestral-7B}).
% ~\cite{dao2024transformersssmsgeneralizedmodels}.
  } 
  \vspace{-0.25in}
  \label{fig:evcal_R2_arch}
\end{figure}

We select three representative \LLMs to cover the main architecture families used in modern language models.
We use \texttt{Qwen3-8B}~\cite{yang2025qwen3technicalreport} as a representative dense transformer model, \texttt{Qwen3-30B-A3B}~\cite{yang2025qwen3technicalreport} as a 30 billion parameter \MoE model with 3 billion parameters activated per token, and \texttt{Mamba-Codestral-7B}~\cite{dao2024transformersssmsgeneralizedmodels} as a representative \SSM.
These models allow us to isolate architectural differences while keeping model scale within a comparable range.
Fig.~\ref{fig:evcal_R2_arch} shows the layer-to-final similarity trends for the three architectures.
Dense transformer models exhibit a gradual and relatively smooth increase in logit similarity with depth, indicating progressive refinement of representations.
In contrast, \MoE models concentrate computation within dynamically selected experts and defer final decision making to deeper layers, which reduces cross-layer alignment and limits early-exit opportunities.
\SSMs further minimize redundancy by tightly coupling sequential state updates across depth, making intermediate representations highly dependent on later transformations and therefore least suitable for early-exit.

% Table~\ref{} reports the corresponding oracle-exit results.
% Overall, dense Transformer models achieve higher oracle skip ratios at comparable accuracy levels, while MoE and \SSM models show more limited early-exit potential, suggesting that architectural efficiency does not directly translate into layer redundancy exploitable by early-exit.

\subsection{Training Recipe}
\label{subsec:training_recipe}

We further examine how pretraining progression and post-training tuning affect a model’s suitability for early-exit in this subsection.
The results indicate that \textit{both continued pretraining and post-training alignment tend to reduce the base model’s inherent ability to support effective early-exit}.

\begin{figure}[t]
  \centering
  \includegraphics[width=.98\linewidth]{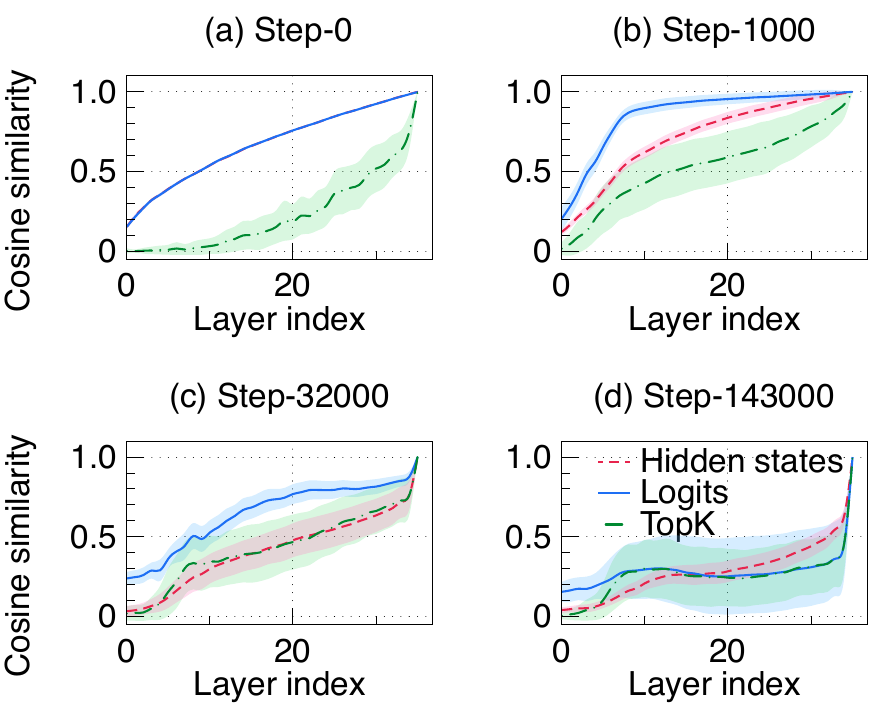}
  \vspace{-0.15in}
  \caption{
The layer-to-final similarity of \texttt{Pythia-12B}
% ~\cite{biderman2023pythia} 
across different training stages.
  } 
  \vspace{-0.2in}
  \label{fig:eval_R2_pretraining}
\end{figure}

\textbf{Pretraining evaluation.}
In fact, most model providers do not release detailed pretraining configurations.
In such a case, we study pretraining effects using models with publicly available training checkpoints.
We analyze Pythia~\cite{biderman2023pythia}, an open-weight transformer model for which intermediate checkpoints throughout pretraining are available.
% This enables us to directly observe how internal layer-to-final similarity evolves as training progresses.
% 
Figure~\ref{fig:eval_R2_pretraining} illustrates the change in layer-to-final similarity across different pretraining stages.
The transition from a linear to a bowed similarity curve indicates functional specialization, where intermediate layers diverging from the final output to perform complex semantic transformations. In mature stages (Step-143000), a low-similarity plateau emerges in the middle layers, which suggests that advanced training concentrates critical decision-making and probability calibration in the final layers. As a result, layer-wise redundancy is reduced, and early-exit becomes less reliable as the pretraining progresses.

\begin{figure}[t]
  \centering
  \includegraphics[width=.98\linewidth]{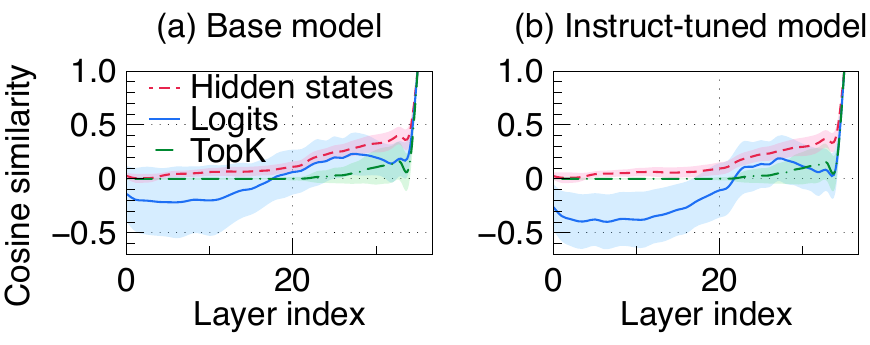}
  \vspace{-0.15in}
  \caption{
The effect of post-training alignments on the model's early-exit adaptability.
  } 
  \vspace{-0.25in}
  \label{fig:eval_R2_posttraiing}
\end{figure}

\textbf{Post-training evaluation.}
To further examine the impact of post-training, we evaluate different variants of the \texttt{Qwen3-4B} model, including the base pretrained model, and an instruction-tuned version.
Figure~\ref{fig:eval_R2_posttraiing} compares the early-exit suitability across these variants.
We observe that post-trained models generally exhibit delayed logit alignment compared to the base model, indicating stronger late-layer calibration.
This behavior is expected because instruction tuning explicitly optimizes the final decoding behavior through supervised objectives.
This suggests that alignment and reasoning-oriented fine-tuning further reduces layer redundancy, making early-exit more challenging in post-trained models.

% \subsection{Inference Setting}

\subsection{Prompt Dataset}

\begin{figure}[t]
  \centering
  \includegraphics[width=.98\linewidth]{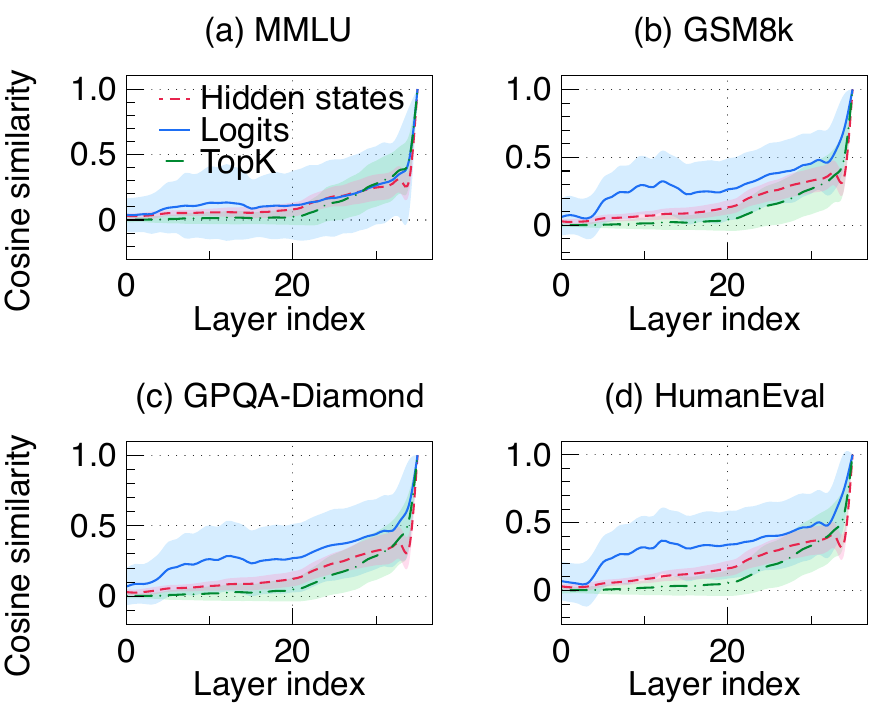}
  \vspace{-0.15in}
  \caption{
The effect of workloads on the model's early-exit adaptability.
  } 
  \vspace{-0.2in}
  \label{fig:eval_R2_dataset}
\end{figure}

Different tasks induce varying output lengths and layer utilization patterns.
To study how workload characteristics affect early-exit behavior, we evaluate the \texttt{Qwen3-8B} model on four representative datasets: MMLU~\cite{hendrycks2021mmlu}, GSM8k~\cite{cobbe2021gsm8k}, GPQA-Diamond~\cite{rein2023gpqagraduatelevelgoogleproofqa}, and HumanEval~\cite{chen2021humaneval}.
Fig.~\ref{fig:eval_R2_dataset} shows a consistent increase in similarity as layer depth grows.
The model shows lower potential on MMLU, as early termination is likely to alter the output. However, GSM8k applies stricter accuracy criteria than MMLU’s multiple-choice evaluation. As a result, the end-to-end benefits on MMLU remain higher than those on GSM8k, as shown in \S\ref{subsec:upper_bound}.

% \subsection{Output token length.}

%% file: sections/related.tex
\section{Related Work}
\label{sec:related}

Prior work shows that representations in transformer models evolve across layers from surface-level features to semantic and task-specific information, while also exhibiting substantial layer-level redundancy~\cite{men-etal-2025-shortgpt,xiao2025densing}. Layer-wise early-exit methods attach auxiliary heads to intermediate layers and terminate inference once an exit criterion is met, sometimes combined with speculative decoding to reduce accuracy loss~\cite{chen2023eellm,pan2024eetuning,xu2025specee,elhoushi-etal-2024-layerskip}. However, existing approaches rely on additional training and are mainly evaluated on earlier \LLM generations such as Llama2~\cite{touvron2023llama2openfoundation}. Our work re-examines early-exit opportunities in modern \LLMs through systematic evaluation.
A detailed discussion of related work is provided in Appendix~\ref{appendix:detailed_related}.

%% file: sections/conclusion.tex
\section{Conclusion}

In this work, we systematically study layer-wise early-exit in modern \LLMs and show that several assumptions behind prior early-exit methods no longer hold as models evolve. 
We introduce an \textit{early-exit adaptability score} and a \textit{unified benchmark} with oracle early-exit evaluation to measure a model’s intrinsic suitability for early-exit and to estimate its upper-bound acceleration potential. This allows researchers to assess the potential benefits of early-exit for a given model and workload before investing effort in designing and implementing specific optimizations.
Our results also reveal a diminishing trend in early-exit adaptability across recent model generations.
At the same time, we identify key factors that shape early-exit potential: although early-exit suitability decreases across model generations, it increases with model scale and is higher for dense transformers. This suggests that early-exit remains promising for large-scale (\eg, more than 20 billion parameters) dense transformers.
% , especially when limited post-training alignment is applied.

%% file: sections/appendix.tex
\appendix

% \section{Detailed Model Selections}
\label{appendix:model}

\section{Selected Model Comparison}
\label{appendix:model_comparison}

We select eight models to analyze layer-to-final similarity in \S\ref{sec:rq1}. Table~\ref{tab:llm_categorization} provides a comprehensive comparison of these model families, including their architectures, key features, and publicly available training recipe information.

\begin{table*}[t]
\centering
\caption{Categorization of Selected \LLM Families.}
\vspace{-0.1in}
\scalebox{0.8}{
\begin{tabular}{lcccc}
\toprule
\textbf{Model Family} & \textbf{Year} & \textbf{Architecture} & \textbf{Key Features} & \textbf{Training Recipe} \\ 
\midrule
Llama2~\cite{touvron2023llama2openfoundation} & 2023 & Dense transformer & RoPE, SwiGLU & 2T tokens, PPO \\ 
Llama3~\cite{grattafiori2024llama3herdmodels} & 2024 & Dense transformer & GQA, 8k Context & 15T tokens, DPO \\ 
Llama4~\cite{meta_llama4_2025} & 2025 & \MoE & Native Multimodal & 10M Context, Reasoning \\ 
\midrule
Qwen2~\cite{yang2024qwen2technicalreport} & 2024 & Dense \& \MoE & YaRN, GQA & 7T tokens, Staged Mix \\ 
Qwen3~\cite{yang2025qwen3technicalreport} & 2025 & Dense \& \MoE & Thinking Mode & CoT-SFT, Multilingual \\ 
\midrule
OpenAI OSS~\cite{openai2025gptoss120bgptoss20bmodel} & 2025 & \MoE & Reasoning MoE & QAT, Configurable Effort \\ 
\midrule
Mamba1~\cite{gu2024mambalineartimesequencemodeling} & 2023 & Selective \SSM & Linear Scan, RNN-like & Hardware-aware Scan \\ 
Mamba2~\cite{dao2024transformersssmsgeneralizedmodels} & 2024 & \SSM-SSD & State Space Duality & MatMul-based (SSD) \\
\bottomrule
\end{tabular}
}
\label{tab:llm_categorization}
\vspace{-0.1in}
\end{table*}

\section{Detailed Early-Exit Adaptability Scores in Recent \LLMs}
\label{appendix:evaluate_ee_score}

We report the detailed \EAS results of popular \LLMs in recent years, with the corresponding model flavor used for evaluation in Table~\ref{tab:model_scores} .

\begin{table}[t]
\centering
\small
\begin{tabular}{lc}
\toprule
\textbf{Model Flavor} & \textbf{Score} \\
\midrule
Llama2-7B~\cite{touvron2023llama2openfoundation} & 0.52 \\
Mamba1-130M~\cite{gu2024mambalineartimesequencemodeling} & 0.35 \\
EE-LLM-7B-dj-refine-150B~\cite{chen2023eellm} & 0.56 \\
Llama3-8B~\cite{grattafiori2024llama3herdmodels} & 0.46 \\
LayerSkip-llama3-8B~\cite{elhoushi-etal-2024-layerskip} & 0.54 \\
Mamba2-7B~\cite{dao2024transformersssmsgeneralizedmodels} & 0.48 \\
Qwen2-7B~\cite{yang2024qwen2technicalreport} & 0.36 \\
Llama3.1-8B~\cite{grattafiori2024llama3herdmodels} & 0.46 \\
Qwen2.5-7B~\cite{yang2024qwen2technicalreport} & 0.41 \\
Llama3.2-8B~\cite{grattafiori2024llama3herdmodels} & 0.45 \\
Llama3.3-8B~\cite{grattafiori2024llama3herdmodels} & 0.44 \\
Gemma3~\cite{gemma_2025} & 0.42 \\
Llama-4-Scout-17B-16E~\cite{meta_llama4_2025} & 0.47 \\
SpecEE (Llama2-7B)~\cite{xu2025specee} & 0.52 \\
Qwen3-8B~\cite{yang2025qwen3technicalreport} & 0.51 \\
Falcon-H1-7B-Base~\cite{falconh1} & 0.44 \\
GPT-OSS-20B~\cite{openai2025gptoss120bgptoss20bmodel} & 0.59 \\
Ministral3-8B-Base~\cite{Mistral3} & 0.44 \\
Nemotron-Cascade-8B~\cite{Nemotron_Cascade} & 0.43 \\
\bottomrule
\end{tabular}
\caption{\EAS for different model flavors used in Fig.\ref{fig:intro}.}
\label{tab:model_scores}
\end{table}

\section{Details of Oracle Early-Exit Strategy}

In \S\ref{subsec:upper_bound}, we study the upper-bound benefits of early-exit methods by evaluating different models under an oracle early-exit setting. Algorithm~\ref{alg:oracle_early_exit} summarizes the oracle early-exit strategy used in our experiments.

\begin{algorithm}[t]
\caption{Oracle Early-Exit Decoding}
\label{alg:oracle_early_exit}
\small
\begin{algorithmic}[1]
\State \textbf{Input:} prompt $p$, threshold $\delta$, max steps $T$
\State \hspace{2.8em} model with $L$ layers and LM head
\State \textbf{Output:} generated text and per-token exit layers

\State $s \gets p$; \ $exit\_layers \gets [\,]$
\For{$t = 1$ to $T$}
    \State Run one forward pass on $s$ and get the last-token hidden state from every layer
    \State Convert each layer's last-token hidden state to logits using the LM head
    \State $z_{\text{final}} \gets$ logits from the last layer
    \State $k^\star \gets L$ \Comment{default: use full depth}

    \For{$k = 1$ to $L$}
        \If{$\mathrm{Sim}(z_k, z_{\text{final}}) \ge \delta$}
            \State $k^\star \gets k$ \Comment{earliest match}
            \State \textbf{break}
        \EndIf
    \EndFor

    \State Next token $\hat{y} \gets \arg\max(z_{k^\star})$
    \State Append $\hat{y}$ to $s$ and record $k^\star$ in $exit\_layers$
    \If{$\hat{y}$ is EOS}
        \State \textbf{break}
    \EndIf
\EndFor
\State \textbf{Output} $(s,\ exit\_layers)$
\end{algorithmic}
\end{algorithm}

\section{Downstream Task Evaluation Setup}
\label{appendix:evaluate}

We use OpenCompass~\cite{2023opencompass}, an open-source \LLM benchmarking system, to run the evaluation experiments mentioned in this paper. In this section, we explain the details of the instructions and the evaluation metrics we use to get the final results.

\subsection{Tuned Instructions}

We demonstrate the tuned instructions used for the inference tasks, including MMLU~\cite{hendrycks2021mmlu}, GSM8k~\cite{cobbe2021gsm8k}, GPQA~\cite{rein2023gpqagraduatelevelgoogleproofqa}, and HumanEval~\cite{chen2021humaneval} in Table~\ref{tab:prompt_templates}.

\begin{table*}[t]
\centering
\small
\begin{tabular}{l p{0.7\textwidth}}
\toprule
\textbf{Dataset} & \textbf{Instruction Template} \\
\midrule
GSM8k &
\texttt{Question: \{question\} \textbackslash\textbackslash{} 
Let's think step by step \textbackslash\textbackslash{} 
Answer:} \\

HumanEval &
\texttt{Read the following function signature and docstring, and fully implement the function described. 
\textbackslash\textbackslash{} Your response should only contain the code for this function. 
\textbackslash\textbackslash{} \{prompt\}} \\

MMLU &
\texttt{Question: \{input\} \textbackslash\textbackslash{} 
A. \{A\} \textbackslash\textbackslash{} 
B. \{B\} \textbackslash\textbackslash{} 
C. \{C\} \textbackslash\textbackslash{} 
D. \{D\} \textbackslash\textbackslash{} 
Answer:} \\

GPQA &
\texttt{Question: \{question\} \textbackslash\textbackslash{} 
A. \{A\} \textbackslash\textbackslash{} 
B. \{B\} \textbackslash\textbackslash{} 
C. \{C\} \textbackslash\textbackslash{} 
D. \{D\} \textbackslash\textbackslash{} 
Answer:} \\
\bottomrule
\end{tabular}
\caption{Prompt templates used for each evaluation dataset.}
\label{tab:prompt_templates}
\end{table*}

\subsection{Accuracy Calculation Rules}

In \S\ref{subsec:upper_bound}, we explore the upper bound benefits that can be brought by early-exit methods, by evaluating varying models using the oracle
early-exit strategy under MMLU~\cite{hendrycks2021mmlu} and GSM8k~\cite{cobbe2021gsm8k} datasets.
In the appendix, we further show how we extract the answer from the output sequences, and compute the final accuracy in Algorithm~\ref{alg:gsm8k_eval} and \ref{alg:mmlu_eval}.

\begin{algorithm}[t]
\caption{GSM8k Evaluation}
\label{alg:gsm8k_eval}
\small
\begin{algorithmic}[1]
\State \textbf{Input:} Predictions $P=\{p_i\}_{i=1}^N$
\State \hspace{2.8em} References $R=\{r_i\}_{i=1}^N$
\State \textbf{Output:} Accuracy score

\If{$|P| \neq |R|$}
    \State \textbf{return} error
\EndIf

\State $correct \gets 0$
\For{$i = 1$ to $N$}
    \State $s_i \gets \mathrm{Split}(p_i,\texttt{``Question:''})[0]$
    \State $N_i \gets \mathrm{RegexFindAll}\big(s_i,\texttt{/-?\textbackslash d+\textbackslash.\textbackslash d+|-?\textbackslash d+/}\big)$
    \State $\hat{a}_i \gets 
    \begin{cases}
    \mathrm{Last}(N_i), & N_i \neq \emptyset \\
    \texttt{NULL}, & \text{otherwise}
    \end{cases}$
    \If{$\hat{a}_i = r_i$ \textbf{or} $\left|\mathrm{float}(\hat{a}_i) - \mathrm{int}(r_i)\right| < \varepsilon$}
        \State $correct \gets correct + 1$
    \EndIf
\EndFor

\State \textbf{Output} $100 \times correct / N$
\end{algorithmic}
\end{algorithm}

\begin{algorithm}[t]
\caption{MMLU Evaluation}
\label{alg:mmlu_eval}
\small
\begin{algorithmic}[1]
\State \textbf{Input:} Predictions $P=\{p_i\}_{i=1}^N$
\State \hspace{2.8em} References $R=\{r_i\}_{i=1}^N$
\State \hspace{2.8em} Prompts $Q=\{q_i\}_{i=1}^N$
\State \textbf{Output:} Accuracy and per-example details

\If{$|P| \neq |R|$}
    \State \textbf{return} error
\EndIf

\State $correct \gets 0$, $total \gets 0$
\State $details \gets \{\}$

\For{$i = 1$ to $N$}
    \State $\hat{y}_i \gets \mathrm{FirstOptionPostprocess}(p_i,\ \{A,B,C,D\})$
    \State $is\_correct \gets (\hat{y}_i = r_i)$
    \If{$is\_correct$}
        \State $correct \gets correct + 1$
    \EndIf
    \State $details[\mathrm{str}(i)] \gets \{\texttt{prompt}: q_i,\ \texttt{pred}: \hat{y}_i,\ \texttt{refr}: r_i,\ \texttt{is\_correct}: is\_correct\}$
    \State $total \gets total + 1$
\EndFor

\State \textbf{return} $\{\texttt{accuracy}: 100 \times correct/total,\ \texttt{details}: details\}$
\end{algorithmic}
\end{algorithm}

% \section{Detailed Similarity Results}

\section{Detailed Related Work}
\label{appendix:detailed_related}

% \noindent\textbf{Conditional and Adaptive Computation in Neural Networks.}
% Conditional computation dynamically adjusts the amount of computation per input to enable the model's abilities on different tasks or reduce inference cost. 
% % 
% This idea has been widely explored in the traditional \ML area, 
% % 
% Early work includes Adaptive Computation Time (ACT), which allows recurrent and transformer models to halt computation once sufficient confidence is reached~\cite{graves2016adaptive}. Subsequent studies explore conditional layer skipping and routing mechanisms to enable input-dependent depth or width~\cite{bengio2015conditional,wang2018skipnet}. Mixture-of-Experts (MoE) architectures further extend this idea by activating only a subset of experts per token, significantly increasing model capacity while keeping per-token computation bounded~\cite{shazeer2017outrageously,fedus2022switch}. These approaches motivate early-exit as a form of adaptive computation, but they do not guarantee that intermediate representations are directly suitable for high-quality generation.

\noindent\textbf{Representation evolution across layers in \LLMs.}
% Previous works have extensively analyzed how representations evolve across transformer layers. 
Previous studies show that lower layers capture lexical and syntactic features, while higher layers encode semantic and task-specific information~\cite{tenney-etal-2019-bert,jawahar-etal-2019-bert}. \citet{ethayarajh-2019-contextual} further shows that representations become increasingly anisotropic and specialized in deeper layers, indicating an opportunity for early-exit for not important tokens (\eg, punctuations). 
Nonetheless, recent studies suggest that modern \LLMs distribute semantic processing more evenly across layers~\cite{yan-2025-addition,yu-etal-2025-back,skean2025layer}.
This indicates a progressive refinement across layers, which leads to frequent output logits and hidden states drifting between layers, potentially reducing the predictive quality of intermediate outputs~\cite{glavas2024dynamiclayerselectiondecoderonly}. These findings motivate a re-examination of whether intermediate layers in recent \LLMs remain suitable for early-exit.

\noindent\textbf{\LLMs' redundancy, sparsity, and density.}
Redundancy is widely observed in nonlinear models, where many components contribute marginally to final predictions~\cite{bian-etal-2021-attention}. In encoder-only models such as BERT~\cite{devlin-etal-2019-bert}, redundancy exists at both representation and neuron levels~\cite{dalvi-etal-2020-analyzing}, and recent work shows that similar layer-level redundancy persists in decoder-only \LLMs~\cite{men-etal-2025-shortgpt,gromov2024unreasonable}. 
To better understand this behavior, recent studies define and measure both activation and layer sparsity in decoder-only models~\cite{luo2024sparsinglaw,huang2025determining,lu2025luallm,csordás2025languagemodelsusedepth}. Meanwhile, \citet{xiao2025densing} demonstrate that the effective density of modern \LLMs continues to increase with scale and evolving optimization techniques, indicating decreasing opportunities for early-exit in advanced \LLMs.

\noindent\textbf{Layer-wise early-exit in \LLMs.}
Layer-wise early-exit attaches auxiliary language model heads to intermediate layers and terminates inference when an exit criterion is met, which has been widely explored in decoder-only \LLMs. 
% 
% Early-exit has been explored in transformer encoders such as BERT through methods like DeeBERT and PABEE, which show latency reductions for classification tasks under negligible accuracy loss~\cite{xin2020deebert,zhou2020pabee}. 
% 
EE-LLM~\cite{chen2023eellm,pan2024eetuning} provides a general framework for large-scale training and inference of early-exit \LLMs.
EESD~\cite{liu-etal-2024-speculative-decoding}, SpecEE~\cite{xu2025specee} and LayerSkip~\cite{elhoushi-etal-2024-layerskip} combine early-exit with speculative decoding~\cite{pmlr-v202-leviathan23a}, aiming to reduce the accuracy degradation.
However, existing work heavily relies on additional training, and has only been evaluated with prior generations of \LLMs, especially Llama2~\cite{touvron2023llama2openfoundation}, which exhibit contrasting behavior compared to the latest models, as shown in \S\ref{sec:rq1}.
In this paper, we evaluate the early-exit opportunities in the latest \LLMs and provide a comprehensive analysis of the causes.
% \subsection{Detailed Early-Exit Adaptability Scores}

\section{Terms of Use and Distribution}

All experiments in this paper use publicly available datasets and open-weight models released by their respective authors. The datasets and model checkpoints are accessed under their original licenses and terms of use. We do not redistribute any datasets or model weights. The code and evaluation scripts that will be released with this work are provided for research and educational purposes only and do not include proprietary content. Users are responsible for ensuring compliance with the licenses and usage policies of the original datasets and models when reproducing our results.